\documentclass[10pt,twocolumn,letterpaper]{article}

\usepackage{wacv}
\usepackage{times}
\usepackage{epsfig}
\usepackage{graphicx}
\usepackage{amsmath}
\usepackage{amssymb}
\usepackage[ruled,vlined]{algorithm2e}
\usepackage{setspace}
\usepackage{tabularx}
\usepackage{makecell}
\usepackage[normalem]{ulem}
\usepackage{graphicx}
\usepackage{comment}
\usepackage{color}
\usepackage{microtype}
\usepackage{subfigure}
\usepackage{booktabs} 

\usepackage{soul}

\newcommand{\hj}[1]{\textcolor{black}{#1}}

\def\wacvPaperID{0084} 

\wacvalgorithmstrack   

\wacvfinalcopy 

\usepackage[pagebackref=true,breaklinks=true,colorlinks,bookmarks=false]{hyperref}

\begin{document}

\title{Improving Multi-fidelity Optimization with a Recurring Learning Rate for Hyperparameter Tuning}

\author{
HyunJae Lee\quad
Gihyeon Lee\quad
Junhwan Kim\quad
Sungjun Cho\quad
Dohyun Kim\quad
Donggeun Yoo\\
Lunit Inc.\\
{\tt\small \{hjlee,gihyeon.lee,kimjh12,scho,donny8,dgyoo\}@lunit.io}
}

\maketitle
\thispagestyle{empty}

\begin{abstract}
Despite the evolution of Convolutional Neural Networks (CNNs), their performance is surprisingly dependent on the choice of hyperparameters. However, it remains challenging to efficiently explore large hyperparameter search space due to the long training times of modern CNNs.
Multi-fidelity optimization enables the exploration of more hyperparameter configurations given budget by early termination of unpromising configurations. However, it often results in selecting a sub-optimal configuration as 
training with the high-performing configuration typically converges slowly in an early phase.
In this paper, we propose Multi-fidelity Optimization with a Recurring Learning rate (MORL) which incorporates CNNs' optimization process into multi-fidelity optimization.
MORL alleviates the problem of slow-starter and achieves a more precise low-fidelity approximation. Our comprehensive experiments on general image classification, transfer learning, and semi-supervised learning demonstrate the effectiveness of MORL over other multi-fidelity optimization methods such as Successive Halving Algorithm (SHA) and Hyperband. Furthermore, it achieves significant performance improvements over hand-tuned hyperparameter configuration within a practical budget.
\end{abstract}

\section{Introduction}
\label{sec:introduction}

Convolutional Neural Networks (CNNs) have recently achieved a huge success in a wide range of computer vision tasks ~\cite{lee2019srm, long2015fully, nam2021reducing, ren2015faster}. 
While the performance of CNNs is greatly affected by the choice of hyperparameters 
\cite{choi2019empirical}, the optimal combination of hyperparameters is hard to know a priori.
Therefore, \hj{model developers often explore} the hyperparameter configurations \hj{}{manually,} which requires \hj{a huge labor cost} yet it is sub-optimal with respect to performance \cite{bergstra2012random}.
Recent approaches in hyperparameter optimization (HPO) \cite{bergstra2011algorithms, li2017hyperband} try to automate this \hj{painful} tuning process by efficiently exploring multi-dimensional search space. In practice, it is known that the automatic tuning process has contributed significantly to improving the \hj{winning} rate during the development of AlphaGo \cite{chen2018bayesian}. 

As modern CNNs get more sophisticated and complex, the search space of hyperparameters becomes larger and the training time becomes longer\hj{, which makes it difficult to explore diverse configurations.}
In order to speed up the HPO process and save computational expense, multi-fidelity optimization early stops unpromising configurations and adaptively allocates more resources to promising configurations.
The early termination of configuration is determined by a low-fidelity approximation based on partial training results which is computationally much cheaper than full training. 

\begin{figure}[t]
\begin{center}
\includegraphics[width=0.47\textwidth]{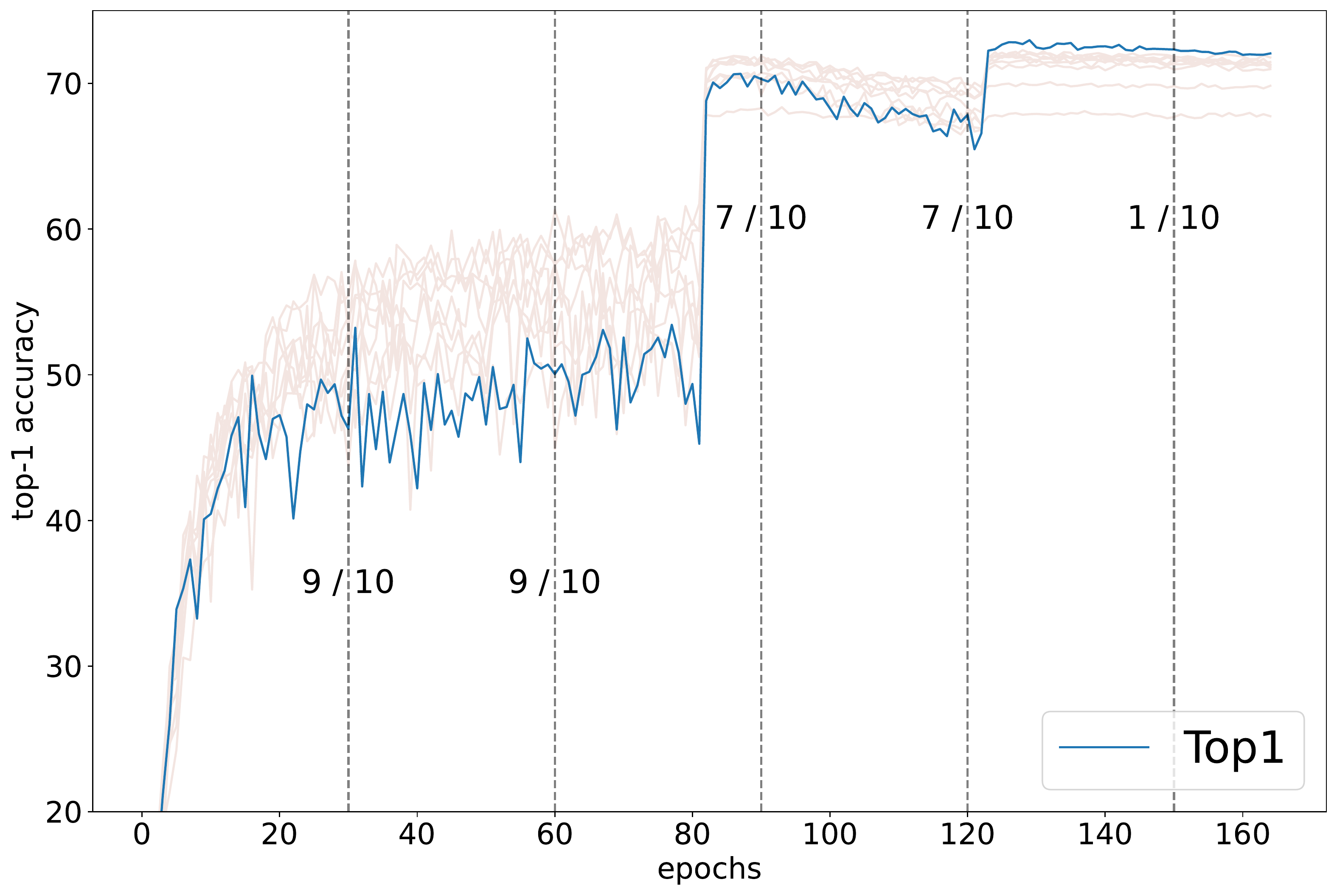}
\caption{Validation curves of grid search on learning rate from 0.01 to 0.1 with ResNet-56 on CIFAR-100 dataset (see section \ref{sec:object-cls} for the details). \hj{The fractional numbers on the plot} show the ranking of the best configuration \hj{(blue curve)} every 30 epochs. The best configurations do not outperform others until the late phase.}
\label{fig:grid_search}
\end{center}
\end{figure}

Successive Halving Algorithm (SHA) \cite{jamieson2016non} and Hyperband \cite{li2017hyperband} are the two most \hj{popular} multi-fidelity optimization methods. These methods explore orders-of-magnitude more configurations by early stopping low-performing configurations.
However, we experimentally discovered that the high\hj{-}performing configurations typically do not perform well in the early phase when training CNNs. 
Figure~\ref{fig:grid_search} illustrates the validation curves for the grid search of learning rate with ResNet-56 on CIFAR-100 dataset. It \hj{demonstrates} that the best configuration does not outperform other configurations until the late phase of the training process.
This slow-starting peculiarity of promising configurations limits the performance of multi-fidelity optimization when training CNNs due to the early termination of promising configurations.

\hj{A CNN is learned with a predefined learning rate schedule in general. From the beginning to the end of the learning rate schedule influences the final performance of the CNN. However, previous methods discard low-performing configurations at an early/mid-point of the learning rate schedule and consequently result in a sub-optimal configuration.}
Inspired by this \hj{problem}, we propose Multi-fidelity Optimization with a Recurring Learning rate (MORL) which condenses the learning rate schedule to fit into every round of promotion. Even at an early phase of training, it enables precise evaluation of a configuration as the model comes from the entire schedule.
To the best of our knowledge, this is the first work that integrates the learning rate schedule into multi-fidelity optimization.

Our experiments on a wide range of general computer vision tasks including image classification \cite{he2016deep}, transfer learning \cite{zamir2018taskonomy} and semi-supervised learning \cite{arazo2020pseudo} verify the efficacy of the proposed method. Throughout the experiments, MORL consistently outperforms other multi-fidelity optimization methods such as SHA and Hyperband. Furthermore, it outperforms highly \hj{hand}-tuned configurations significantly within a reasonable budget. 
\hj{Also, we show that our} method is orthogonal to the existing Bayesian optimization method and applicable to a variety of learning rate schedules. \hj{The experimental results demonstrate that MORL is a practical method available for} a wide range of scenarios.

\section{Related Work}
\label{sec:related_work}

\paragraph{Hyperparameter Optimization}
\hj{A CNN} includes many hyperparameters and \hj{it is very crucial to find a good hyperparameter configuration to achieve successful performance}~\cite{bergstra2012random}. While the search space for hyperparameter is often very high-dimensional and complex, it is hard to apply classical optimization methods \hj{that incorporate} gradient descent, convexity or smoothness \cite{feurer2019hyperparameter}.  Therefore, \hj{model developers} often go through a manual tuning process which requires \hj{a huge labor cost} \cite{bergstra2012random}. Hyperparameter optimization (HPO) aims to automate this \hj{painful} process and search good hyperparameter configurations efficiently.

Bayesian optimization \cite{bergstra2011algorithms, chen2018bayesian} optimizes black-box function by adaptively suggesting hyperparameter configurations given observations. In order to estimate the target function, it first fits a probabilistic surrogate model with observations of input configuration and its corresponding performance. Then it selects the next configuration that maximizes acquisition function\hj{. For instance, expected improvement \cite{jones1998efficient} attempts to trade off exploration against exploitation.}
By iteratively fitting a surrogate model and evaluating probable configuration, Bayesian optimization outperforms brute-force methods such as random search \cite{bergstra2011algorithms, hutter2011sequential}. However, Bayesian optimization inherently can not be easily parallelized due to the nature of sequentially fitting a probabilistic model with previous observations \cite{li2020system}. 
Furthermore, it is known that Bayesian optimization does not adapt well to high-dimensional search spaces where it shows similar performance to random search \cite{wang2013bayesian}.

More related to our \hj{method}, multi-fidelity optimization \hj{methods enable to evaluate} order-of-magnitude more configurations by exploiting cheaper proxy tasks, e.g. training model only for few iterations, using partial datasets or downsized images \cite{bergstra2011algorithms, li2017hyperband, klein2017fast}.
As training a single configuration could take from days to weeks due to the increasing model \hj{complexity} and dataset size \cite{xie2020self}, it becomes more crucial to speed up the HPO process by harnessing the power of multi-fidelity optimization.
Successive Halving Algorithm (SHA) \cite{jamieson2016non} uniformly allocates a small initial budget among randomly selected configurations and early stops the worst half configurations. It then doubles the budget and repeats the same process until it reaches maximum resource. \hj{In addition to} its simplicity, SHA shows comparable performance \hj{to} other state-of-the-art HPO methods such as Vizier \cite{golovin2017google}, FABOLAS \cite{klein2017fast} and PBT \cite{jaderberg2017population} in a wide range of tasks \cite{li2020system}.
Hyperband further extends SHA by automating the choice of initial budget via running different variants of SHA with respect to initial resource. BOHB \cite{falkner2018bohb} combines Hyperband with Bayesian optimization in order to benefit from both adaptive resource allocation and configuration sampling.
Our approach enables more precise early stage ranking among configurations by incorporating the learning rate schedule into multi-fidelity optimization.

\paragraph{Recurring learning rate schedule}
Learning rate is one of the most important factors when tuning the performance of CNNs \cite{smith2017cyclical}. 
A typical learning rate schedule used by a wide range of modern CNNs is a step learning rate schedule where an initial learning rate is decayed by an adequate factor at given milestone epochs \cite{krizhevsky2012imagenet, he2016deep, simonyan2014very, huang2017densely, szegedy2015going}.
In an attempt to eliminate the need for tuning the initial learning rate and its schedule, Cyclical Learning Rates (CLR) \cite{smith2017cyclical} monotonically increases then decreases the learning rate within reasonable bounds and repeats this process cyclically. 
Stochastic gradient descent with warm restart (SGDR) \cite{loshchilov2017sgdr} periodically warm restarts SGD process where the learning rate is reinitialized in each cycle and scheduled to decrease following cosine annealing schedule. SGDR accelerates the training process of CNN and it achieved state-of-the-art performance by an ensemble of models obtained at the end of each cycle.
\hj{Our method differs from these methods in that} we aim to improve multi-fidelity optimization by leveraging the learning rate schedule.

\section{Multi-fidelity Optimization with Recurring Learning Rate}
\label{sec:method}

In this section, we introduce Multi-fidelity Optimization with \hj{a} Recurring Learning rate (MORL) algorithm which extends the Successive Halving Algorithm (SHA) to improve low fidelity approximations by taking the optimization process of CNNs into consideration. We first provide a brief introduction to SHA, motivate the need for a recurring learning rate then introduce MORL algorithm in detail.  

\subsection{Successive Halving Algorithm}
Given a reduction factor $\eta$ and maximum resource $r$, SHA trains the network with an initial resource allocated to each configuration, evaluates the performance of all configurations\hj{,} and promotes top $1 / \eta$ configurations to the next round. It then increases the resource allocation for each configuration by a factor of $\eta$ and repeats the process until the resource allocation per each configuration reaches $r$. 
By allocating relatively small initial resources and early stopping, SHA enables the evaluation of order-of-magnitude more configurations given a fixed budget. However, when low-fidelity approximation does not reflect the final performance, SHA ends up terminating configuration which will have high performance at the end.

Figure \ref{fig:grid_search} depicts the validation curves of grid search on learning rate with stochastic gradient descent (SGD) and step learning rate schedule which is typically employed when training modern CNNs \cite{krizhevsky2012imagenet, he2016deep, simonyan2014very, huang2017densely, szegedy2015going}. 
The top-performing configuration rather shows inferior performance in the early phase but begins to exhibit high performance in the late phase. 
In this scenario, utilizing SHA results in the early termination of the top-performing configuration. While assigning a large initial budget could mitigate this problem, it leads to evaluating only a small number of configurations which limits the benefit of utilizing multi-fidelity optimization.

\begin{figure}
\begin{center}
    \includegraphics[width=0.47\textwidth]{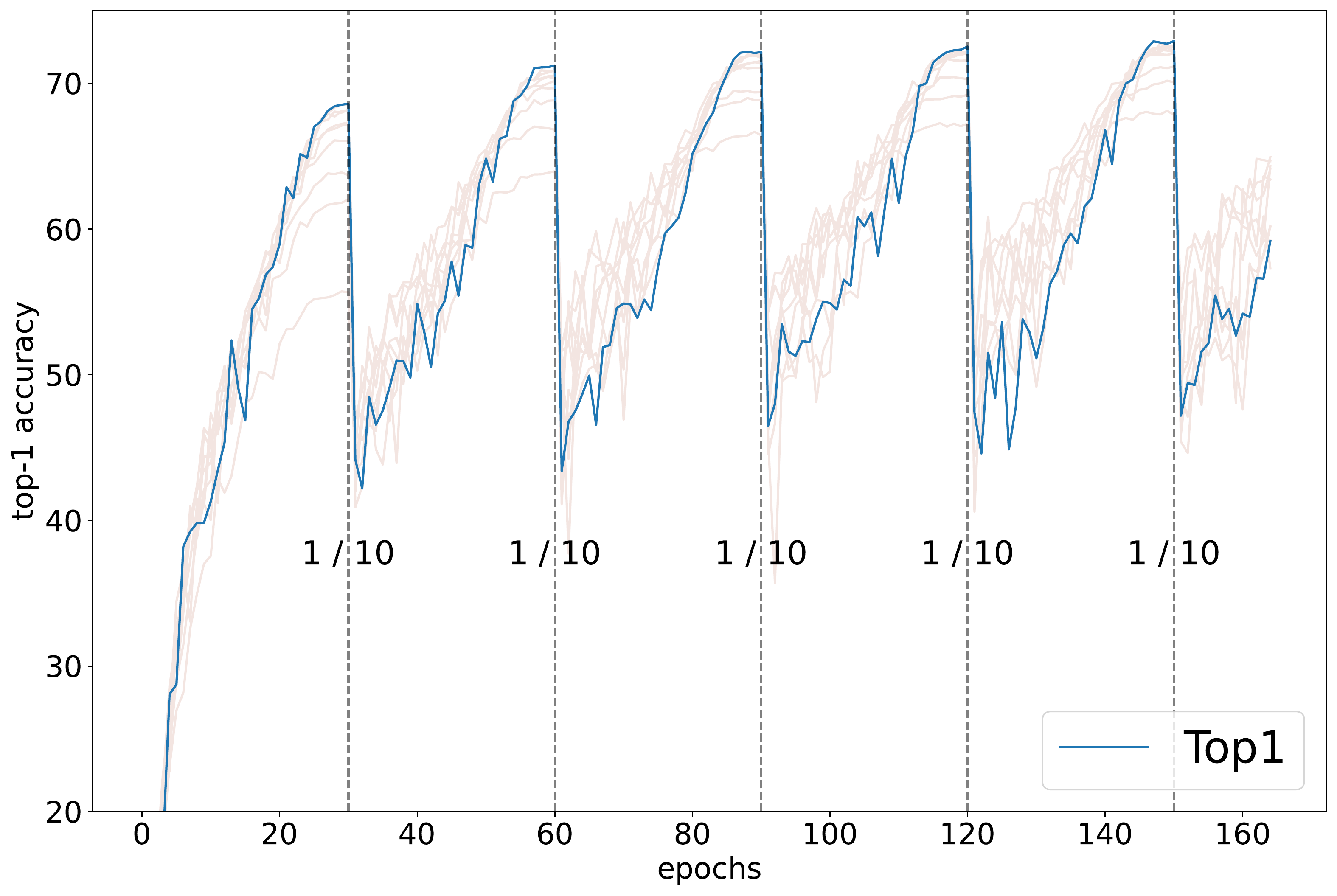}
    \caption{
    Validation curves of grid search on learning rate with a recurring learning rate schedule every 30 epochs, following the same procedure as Figure \ref{fig:grid_search}. The best configuration consistently shows a prominent ranking at the end of each cycle. }
    \label{fig:cawr_grid_search}
\end{center}
\end{figure}

\begin{figure*}[t]
\begin{center}
\includegraphics[width=0.84\textwidth]{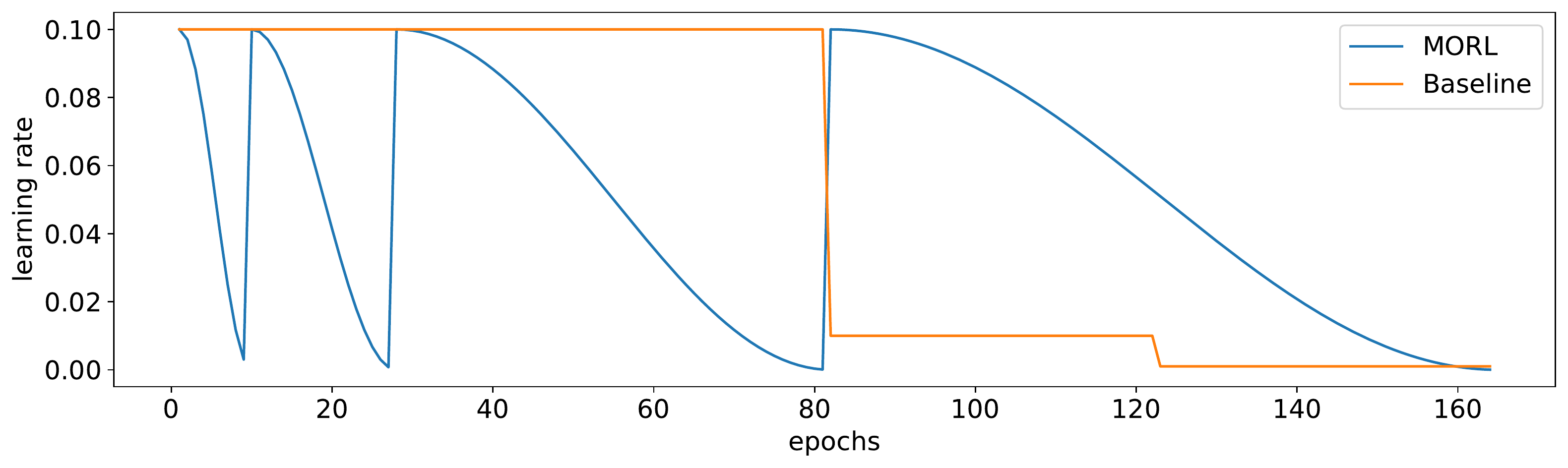}

\caption{Illustration of learning rate schedules utilized in object classification task with maximum resource $r$ of 164 epochs\hj{. The} initial learning rate $l$ \hj{is} 0.1. For the baseline \hj{(orange curve)}, we follow the schedule used in the original implementation.}
\label{fig:lr_schedule}
\end{center}
\end{figure*}

\subsection{Effect of Recurring Learning Rate}
SGD updates the parameter of the network as $\theta = \theta - l g$ given a learning rate $l$, model parameter $\theta$ and gradient $g$. 
Since convergence towards the final performance generally occurs when the magnitude of the parameter update is small \cite{zeiler2012adadelta}, we assumed that it would be more reasonable to evaluate low-fidelity approximation when the learning rate is small.
As a typical learning rate schedule starts with a relatively high learning rate \hj{and} ends with a relatively low learning rate, we condense the original learning rate schedule to fit each round of promotion so that each round ends with a small learning rate.
We refer to this scheme as a recurring learning rate where the learning rate schedule is condensed and restarted every round of promotion.
Figure \ref{fig:cawr_grid_search} follows the same procedure as Figure \ref{fig:grid_search} using a recurring learning rate every 30 epochs with a cosine annealing schedule, and reports the rank of the best configuration at the end of each cycle. When compared with Figure \ref{fig:grid_search}, the slow-starting property of the best configuration is much alleviated while the final performance remains. The top-performing configuration begins to outperform other configurations at the late phase of each cycle where it has a relatively small learning rate which confirms our assumption.

\begin{algorithm}[t]
\setstretch{1.1}
\SetAlgoLined
\textbf{Input:} number of configurations $n$; 
maximum resource $r$;
initial learning rate $l$;
reduction factor $\eta$ (default : 3);
minimum exponent $s_{min}$ (default : 2); 
\\
$H=$\texttt{suggest\_hyperparameters}$(n)$\\
\For{$h \in H$}{ 
    $G_h = $ \texttt{create\_network($h$)}
}
\For{$s \in \{s_{min}, s_{min}+1, \ldots, \lfloor \log_\eta(r)\rfloor\}$}{
    $e_{start}$ = \texttt{get\_start\_epoch($s$)} \\
    $e_{end}$ = \texttt{get\_end\_epoch($s, r$)} \\
    \For{$h \in H$}{ 
        \For{$e \in \{e_{start}, e_{start}+1, \ldots, e_{end}\}$}{
            $l_{e} = \frac  {l} {2} (1 + \cos{\frac {e\pi} {e_{end} - e_{start}}})$ \\
            \texttt{train\_an\_epoch($G_h, l_{e}, h$)}
        }
        $P_h = $ \texttt{compute\_performance($G_h, h$)}
    }
	$H=$\texttt{top\_k}$(H,P, \eta )$ \texttt{// promote top $1 / \eta$ configurations to next round}
}
\caption{MORL Algorithm.}
\label{algo:morl}
\end{algorithm}

\subsection{MORL Algorithm}
Among various multi-fidelity optimization methods, we opt for SHA due to its simplicity and theoretically principled justification. While maintaining the essence of SHA, MORL enhances the ability to early differentiate promising configurations with a recurring learning rate.
The overall procedure of MORL is summarized in Algorithm \ref{algo:morl}.
MORL first suggests a set of hyperparameter configuration $H$ with \texttt{suggest\_hyperparameters} subroutine given a number of configurations $n$, then it initializes the network with \texttt{create\_network} subroutine for each hyperparameter $h$.
There exists a total of $\lfloor \log_\eta(r)\rfloor\ - s_{min} + 1$ rounds of promotion where minimum resource of $\eta^{{s_{min}}}$ is allocated to each configuration on the first round and increases by a factor of $\eta$ every round until it reaches maximum resource $r$.

Each \texttt{get\_start\_epoch} and \texttt{get\_end\_epoch} subroutine computes the beginning and last epoch given exponent $s$ in each round, respectively. The start epoch $e_{start}$ is set to $\eta^{s-1} + 1$, except for the first round where it is set to $1$. On the other hand, the end epoch $e_{end}$ is set to $\eta^s$ except for the last round which is set to $r$.
In each round, every hyperparameter $h \in H$ is evaluated after training with allocated resources.
Given an initial learning rate $l$, the starting learning rate for each epoch $l_e$ is calculated with a cosine annealing schedule \cite{loshchilov2017sgdr} as $l_{e} = \frac {l} {2} (1 + \cos{ {e\pi} / ({e_{end} - e_{start})}})$.
While hyperparameter $h$ might include \hj{the} initial learning rate $l$, we notate it independently for simplicity of the algorithm. 
\texttt{train\_an\_epoch} subroutine trains the network $G_h$ for an epoch with hyperparameter $h$ and learning rate $l_e$ which is updated after every gradient descent following a cosine annealing schedule.
Finally, the performance for each configuration is computed then top $1 / \eta$ configurations are promoted to the next round. We opt for a cosine annealing schedule due to its nature of recurrence. However, it is worth noting that the effectiveness of MORL is not restricted to certain learning rate schedules but can be combined with various learning rate schedules\hj{. This is} demonstrated in section \ref{sec:various_lr}.

The value of reduction factor $\eta$ and minimum exponent $s_{min}$ adjust the number of promotion rounds and minimum resource allocation.
Since top ${1} / {\eta}$ configurations are promoted and resource increases by a factor of $\eta$ on each round, the larger $\eta$ leads to more aggressive termination and fewer rounds of promotion.  
We set default $\eta$ as 3 because given certain conditions, it is theoretically optimal to set $\eta = e \approx 3$ \cite{li2017hyperband}. Nevertheless, we empirically found that MORL is quite robust to the choice of $\eta$ and generally works well with the value of 2, 3\hj{,} and 4.

In terms of minimum exponent $s_{min}$, there exists a trade-off between the number of configurations $n$ and minimum resource $\eta^{s_{min}}$ allocated to each configuration given a fixed budget. While small $s_{min}$ allows exploring more configurations, it might prematurely evaluate the performance. On the other hand, large $s_{min}$ allows a more precise low-fidelity approximation but only a small number of configurations can be evaluated.
Following the empirical analysis of Cyclic LR \cite{smith2017cyclical} that suggests the length of cycle to be between 4 to 20 epochs, we set minimum exponent $s_{min}=2$ which sets the minimal resource $\eta ^ {s_{min}} = 9$. Our ablation study on $s_{min}$ in section \ref{sec:ablation} further verifies the validity of our setting.

\section{Experiments}
\label{sec:experiments}

\begin{figure*}[t]
\begin{center}
\
\subfigure[VGG11]{
\label{fig:vgg11}
\includegraphics[width=0.45\textwidth]{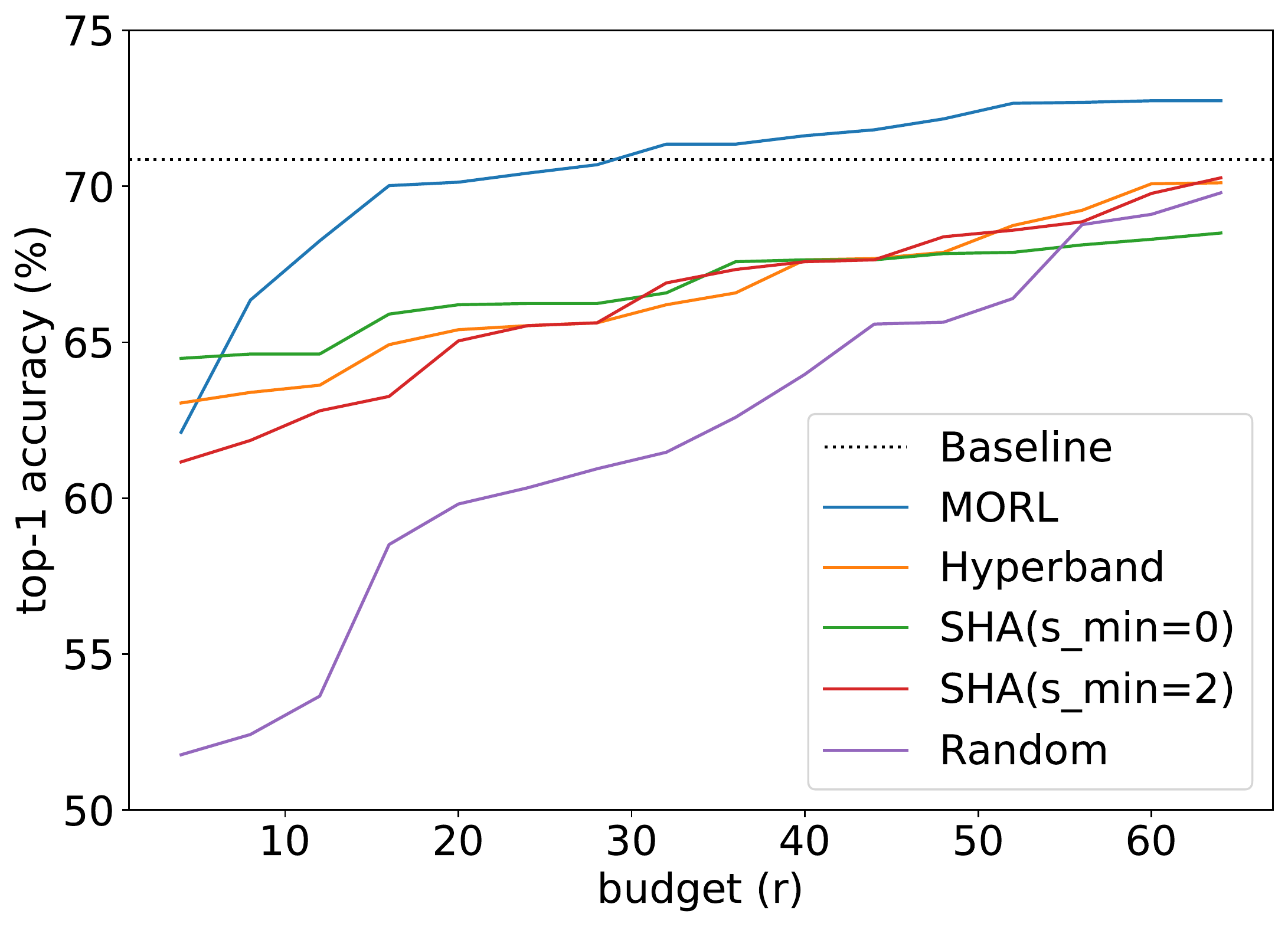}
}
\
\subfigure[AlexNet]{
\label{fig:alexnet}
\includegraphics[width=0.45\textwidth]{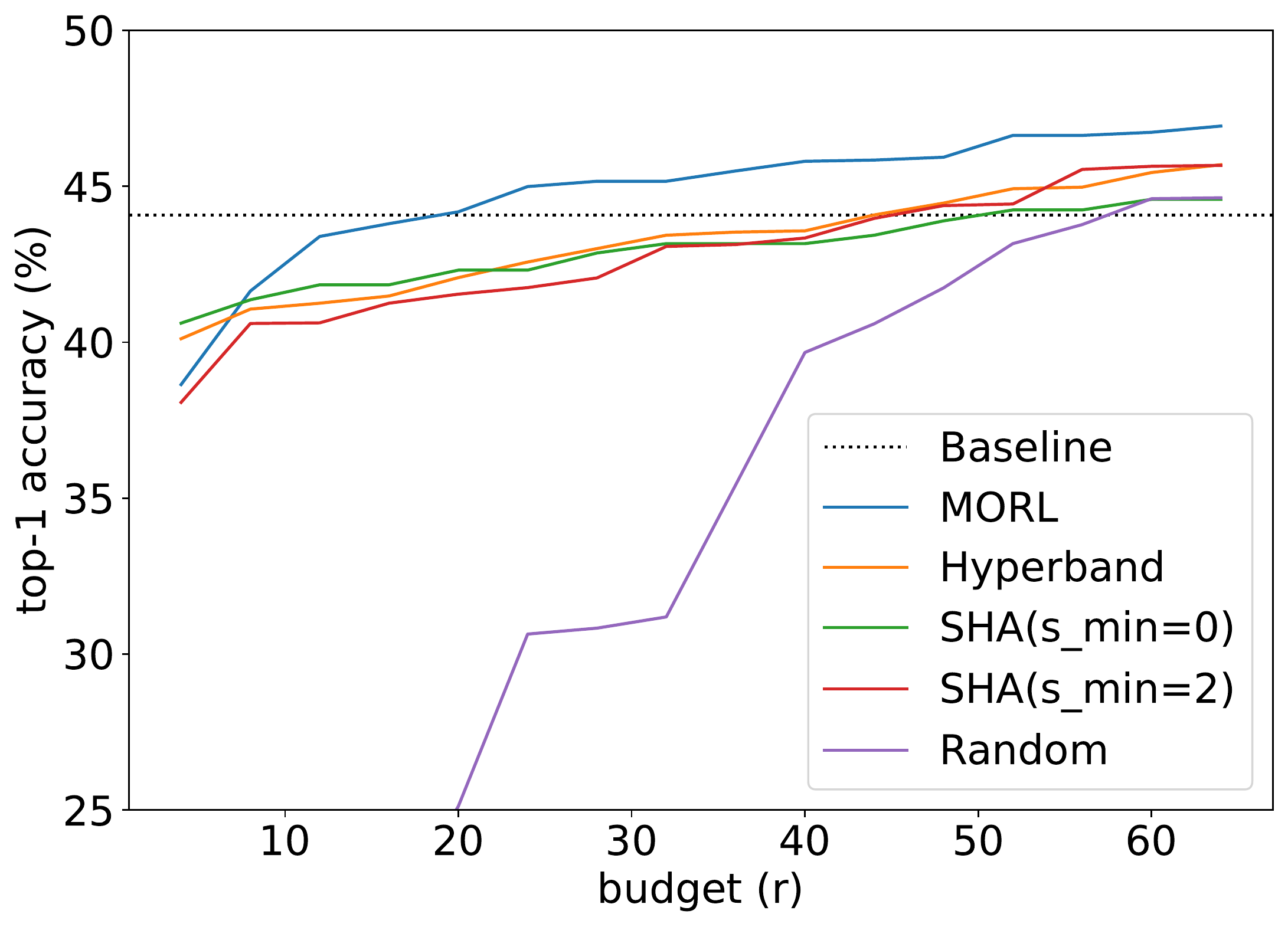}
}
\
\caption{Performance of \hj{various} multi-fidelity optimization methods on CIFAR-100 with (a) VGG-11 and (b) AlexNet architecture. Throughout the optimization process, MORL significantly outperforms other methods except for the very early phase.}
\label{fig:method_comparison}
\end{center}
\end{figure*}

In this section, we conduct a comprehensive evaluation on a wide range of computer vision tasks including object classification, transfer learning, and semi-supervised learning to verify the effectiveness of MORL for hyperparameter optimization. We focus on comparison with SHA \cite{jamieson2016non} and Hyperband \cite{li2017hyperband} which are the de-facto standard multi-fidelity optimization methods implemented in widely adopted HPO frameworks \cite{optuna_2019, liaw2018tune}. 

Our search space consists of learning rate $l$, weight decay $w$, momentum $1 - m$ and batch size $b$ where $l, w$ are sampled from log [$10^{-6}, 10$], $m$ from log [$10^{-6}, 1$], and $b$ from [16, 256].
While there exist various tasks that include task-specific hyperparameters, we believe our search space would serve as a good starting point that could be applied to a wide range of tasks.
\hj{For all} experiments, one unit of resource corresponds to one epoch and the maximum resource $r$ is set to the training epochs \hj{specified} in the original implementation \hj{of each task}. For each \hj{HPO experiment}, we allocate a budget of $64r$ which corresponds to 64 different runs of an experiment when early-stopping is not applied. 

\subsection{Object Classification}
\label{sec:object-cls}

We first evaluate MORL on object classification with publicly available implementation\footnote{https://github.com/bearpaw/pytorch-classification} of classification networks and training schemes for a fair comparison with the human baseline. 
For the baseline, we use the hyperparameter configuration suggested in the original implementation. 
We adopt three different datasets: CIFAR-10/100 datasets \cite{krizhevsky2009learning} \hj{comprising} 50,000 training and 10,000 test images of \hj{10/100 object classes} and Tiny ImageNet dataset  \cite{russakovsky2015imagenet} \hj{containing} 100,000 training and 10,000 validation images of 200 \hj{object} classes. While the original images of Tiny ImageNet consist of 64\hj{$\times$}64 pixels, they are downsized to 32\hj{$\times$}32 pixels for the consistency with CIFAR dataset training process.

We follow the standard practice for data augmentation \cite{he2016deep} where each image is zero-padded with 4 pixels then randomly cropped to the original size and evaluation is performed on the original images. 
We utilize SGD optimizer with an initial learning rate of 0.1, a weight decay of 5e-4, a momentum of 0.9 and a batch size of 128 on a single GPU for the baseline configuration. 
Following the original implementation, the networks are trained for 164 epochs and a step learning rate schedule is applied where the initial learning rate is divided by 10 at 81 and 122 epochs. The illustration of the learning rate schedule for the baseline and MORL is shown in Figure \ref{fig:lr_schedule}.

\textbf{Comparison of multi-fidelity optimization methods.}
We first evaluate MORL compared to other multi-fidelity optimization methods with varying initial resources.
Figure \ref{fig:method_comparison} illustrates the performance of multi-fidelity optimization methods along with the baseline and random search on CIFAR-100 with VGG-11 \cite{simonyan2014very} and AlexNet \cite{krizhevsky2012imagenet} architecture. 
Each SHA with $s_{min}=0$ and $s_{min}=2$ represents the most and intermediate aggressive method of SHA, Hyperband performs grid search over possible values of $s_{min}$, 
and random search performs full training without early-stopping. Throughout the optimization process except for the very beginning, MORL exhibits considerably higher performance compared to other methods. 
In the very early stage, SHA($s_{min}=0$) and Hyperband achieve moderate performance by exploring more configurations and early-stopping hopeless configurations which usually exhibit near random-guessing accuracy.
However, they \hj{slowly improve the performance} whereas MORL improves at a rapid pace.
For the rest of the experiments, we utilize SHA($s_{min}=2$) for the SHA method which exhibits better characteristics than SHA($s_{min}=0$).

\begin{table}
\addtolength{\tabcolsep}{-1pt}
    \caption{Top-1 accuracy (\%) on CIFAR10/100 and Tiny-ImageNet with \hj{VGG-11}. The reported results are the average and the 95\% confidence interval over 5 repetitions. MORL outperforms hand-tuned baseline and other competing methods across varying datasets, and obtains relatively narrow confidence interval. }
\begin{center}
\begin{tabular}{|c|c|c|c|c|c}
\hline
 & CIFAR-10 & CIFAR-100 & Tiny-ImgNet \\
\hline
Baseline & 91.89+-0.25 & 70.89+-0.26 & 48.21+-0.44 \\ 
Random & 91.65+-0.59 & 69.79+-1.84 & 47.63+-2.66 \\ 
SHA & 91.37+-0.35 & 70.27+-1.47 & 48.34+-1.95  \\ 
Hyperband & 91.46+-0.52 & 70.11+-1.09 & 47.99+-1.75  \\ 
MORL & \textbf{92.94+-0.16} & \textbf{72.74+-0.42} & \textbf{50.96+-0.33}  \\ 
\hline
\end{tabular}
\end{center}
\label{table:various_datasets}
\end{table}

\textbf{Scalability to different datasets.} 
We demonstrate the effectiveness of MORL with respect to various datasets of CIFAR-10/100 and Tiny ImageNet. Table \ref{table:various_datasets} compares the top-1 accuracy with the average and the 95\% confidence interval over 5 repetitions. MORL outperforms all other methods in all of the datasets. Whereas other multi-fidelity optimization methods often fail to surpass the hand-tuned baseline given a limited budget and even exhibit lower performance than random search, MORL consistently boosts the performance of the baseline by a \hj{significant} margin. 
It is worth noting that the model acquired by MORL does not introduce any additional weights or computational overhead. The performance gain merely comes from tuning hyperparameters of CNN's optimization process.

\textbf{Scalability to various CNN architectures.} We further verify the scalability of MORL \hj{with respect to} diverse CNN architectures \hj{including} AlexNet, ResNet-20, ResNet-56 and VGG-16 on the CIFAR-100 dataset.
As shown in Table \ref{table:various_archs}, MORL consistently outperforms other methods and improves the performance of the baseline in all experiments. Our results demonstrate the efficacy of MORL with respect to varying CNN architectures.
While SHA and Hyperband only succeed to improve baseline performance on AlexNet, MORL brings meaningful performance improvement even for the relatively large model such as VGG-16.
It further implies the importance of HPO that the performance gain obtained by tuning hyperparameter is not simply substituted by increasing the capacity of CNN architecture. 

\begin{table}
\addtolength{\tabcolsep}{-2.3pt}
\caption{Top-1 accuracy (\%) on CIFAR-100 with different network architectures. MORL consistently improves the performance of baseline with a meaningful margin in all tested networks.}
\begin{center}
\begin{tabular}{|c|c|c|c|c|c}
\hline
  & AlexNet & ResNet-20 & ResNet-56 & VGG-16 \\
\hline
Baseline & 44.08 & 68.67 & 71.54 & 73.58 \\ 
Random & 44.63 & 66.86 & 69.75 & 73.84 \\ 
SHA & 45.67 & 67.15 & 70.41 & 72.48 \\ 
Hyperband & 45.69 & 67.31 & 71.03  & 72.66 \\ 
MORL & \textbf{46.92} & \textbf{69.51} & \textbf{72.87} & \textbf{76.01}  \\ 
\hline
\end{tabular}
\end{center}
\label{table:various_archs}
\end{table}

\subsection{Transfer Learning}
\label{sec:transfer-learning}

A wide range of computer vision tasks employs a pre-trained model that is trained on a large-scale dataset such as ImageNet \cite{russakovsky2015imagenet}. By utilizing meaningful features learned from a large amount of data, transfer learning has successfully boosted the performance of CNNs in various tasks \cite{tan2018survey}. 
In this scenario, it is reasonable to suspect that the problem of slow-starter \hj{is probably} mitigated \hj{as the network converges fast from the pre-trained weights.}

We follow the same training policy as our object classification experiment except for the initial learning rate of the baseline which is reduced by a factor of 10, following common practice for transfer learning. We opt for the ImageNet pre-trained VGG-16 network, which is widely adopted in various computer vision tasks \cite{garg2016unsupervised, long2015fully, ren2015faster} and train the network on CIFAR-10/100 and Tiny ImageNet dataset. We report the top-1 validation accuracy in Table \ref{table:transfer_learning}. 
As expected, SHA and Hyperband show improved performance with respect to the baseline, whereas they often failed to outperform when trained from scratch \hj{as shown in Table~\ref{table:various_datasets}~and~\ref{table:various_archs}}.
Nevertheless, MORL consistently outperforms other methods by a meaningful margin which suggests that the slow-starting tendency is mitigated but remained in the transfer learning setting as well. 

\begin{table}
\caption{Top-1 accuracy (\%) of transfer learning on CIFAR-10/100 and Tiny-ImageNet with VGG-16 pretrained on ImageNet dataset.}
\begin{center}
\begin{tabular}{|c|c|c|c|c|c}
\hline
  & CIFAR-10 & CIFAR-100 & Tiny-ImgNet \\
\hline
Baseline & 93.79 & 74.97 & 53.26 \\ 
Random & 93.50 & 74.47 & 52.64 \\ 
SHA & 94.17 & 75.22 & 54.11  \\ 
Hyperband & 94.01 & 74.99 & 53.96  \\ 
MORL  & \textbf{94.46} & \textbf{76.86} & \textbf{55.52}  \\ 
\hline
\end{tabular}
\end{center}
\label{table:transfer_learning}
\end{table}

\subsection{Semi-Supervised Learning}
\label{sec:semi-supervised-learning}

We finally demonstrate the effectiveness of MORL in semi-supervised learning (SSL) task which jointly learns from a small number of labeled samples and a large number of unlabeled samples. 
Since considerable amounts of unlabeled images are utilized with the supervision from scarce labeled images, \hj{it is crucial to find a good hyperparameter configuration}.
We make use of pseudo-label based SSL algorithm \cite{arazo2020pseudo} which outperforms state-of-the-art consistency regularization methods. For a fair comparison, we adopt their official implementation\footnote{https://github.com/EricArazo/PseudoLabeling} utilizing its default training seed \hj{in order to} compare \hj{different methods} under consistent settings and facilitate further research.

The hand-tuned baseline consists of two-stage training\hj{. It} first trains the network only with the labeled data as a warm-up phase, then finetunes the network with both labeled and unlabeled data.
While the warm-up stage \hj{stabilizes} the training process by initializing unlabeled data with more reliable prediction, it incurs \hj{model developers} to explore a much larger search space since the hyperparameter for each stage needs to be tuned independently. Furthermore, it requires additional resources to train a model \hj{for the first stage. However, if HPO provides better configurations which enable stable training of the second stage, we can omit the warm-up stage.}
Therefore, we exclude the warm-up stage and train the network from scratch for multi-fidelity optimization methods. However, for the baseline, we follow the original two-stage training as the author suggested.

We utilize 13-CNN architecture \cite{athiwaratkun2018there} which is mainly explored in the paper and follow its data processing and optimization setting. Following the original implementation, the network is trained using SGD optimizer with Dropout \cite{srivastava2014dropout} and Mix-up \cite{zhang2017mixup} regularization for 400 epochs where the learning rate is decayed by a factor of 10 at 250 and 350 epoch for the baseline. Table \ref{table:semi_supervised} shows the top-1 accuracy of CIFAR-100 dataset with respect to a varying number of labeled samples.
Surprisingly, MORL achieves significant performance improvement over the baseline throughout the experiments even without the warm-up stage. It further demonstrates the efficacy of MORL in various settings where it works well with a varying number of labeled samples and strong regularization methods such as Mix-up and Dropout.

We further report the top-performing configuration for varying number of labels in Table \ref{table:ssl_hyperparam}, where each setting shows a distinct configuration. While it remains challenging to perform precise analysis due to the nature of hyperparameters, we observe that higher weight decay is applied when there exists a small number of labeled images. It conforms to common intuition that more regularization is desired when a small number of data is given as the network tends to overfit. Our results suggest that it is important to tailor hyperparameters in each experiment setting, a process that can be effectively automated by MORL.

\begin{table}
\caption{Top-1 accuracy (\%) of semi-supervised learning on CIFAR-100 with respect to a varying number of labeled data. While MORL does not include a warm-up stage, it significantly boosts the performance of the baseline which consists of two-stage training. }
\begin{center}
\begin{tabular}{|c|c|c|c|c|c}
\hline
 & 500 labels & 1000 labels & 2000 labels \\
\hline
Baseline & 29.73 & 45.71 & 55.56 \\ 
Random & 31.19 & 46.65 & 54.08   \\ 
SHA & 32.77 & 45.21 & 54.65   \\ 
Hyperband & 32.56 & 45.51 & 54.99   \\ 
MORL & \textbf{35.22} & \textbf{48.52} & \textbf{57.18}  \\ 
\hline
\end{tabular}
\end{center}
\label{table:semi_supervised}
\end{table}

\begin{table}
\caption{The best configuration obtained with MORL algorithm in semi-supervised learning experiments. \hj{The optimized WD value decreases as it uses more labeled samples.} (LR: learning rate, WD: weight decay, MMT: momentum, BS: batch size)}
\begin{center}
\begin{tabular}{|c|c|c|c|c|c|c|}

\hline
 & LR & WD & MMT & BS  \\
\hline
500 labels & 0.0242 & 0.0043 & 0.5652  & 92 \\
1000 labels & 0.0311 & 0.0032 & 0.6834  & 87 \\
2000 labels & 0.0238 & 0.0027 & 0.3926  & 84 \\

\hline
\end{tabular}
\end{center}
\label{table:ssl_hyperparam}
\end{table}

\section{Ablation Study and Analysis}
\label{sec:analysis}
In this section, we perform an ablation study and analytical experiments to gain an insight into the algorithm design choice and potential extensions of MORL. 
We follow the same procedure of object classification experiment on CIFAR-100 in section \ref{sec:object-cls} for the empirical study. 

\begin{figure}
\begin{center}
    \includegraphics[width=0.45\textwidth]{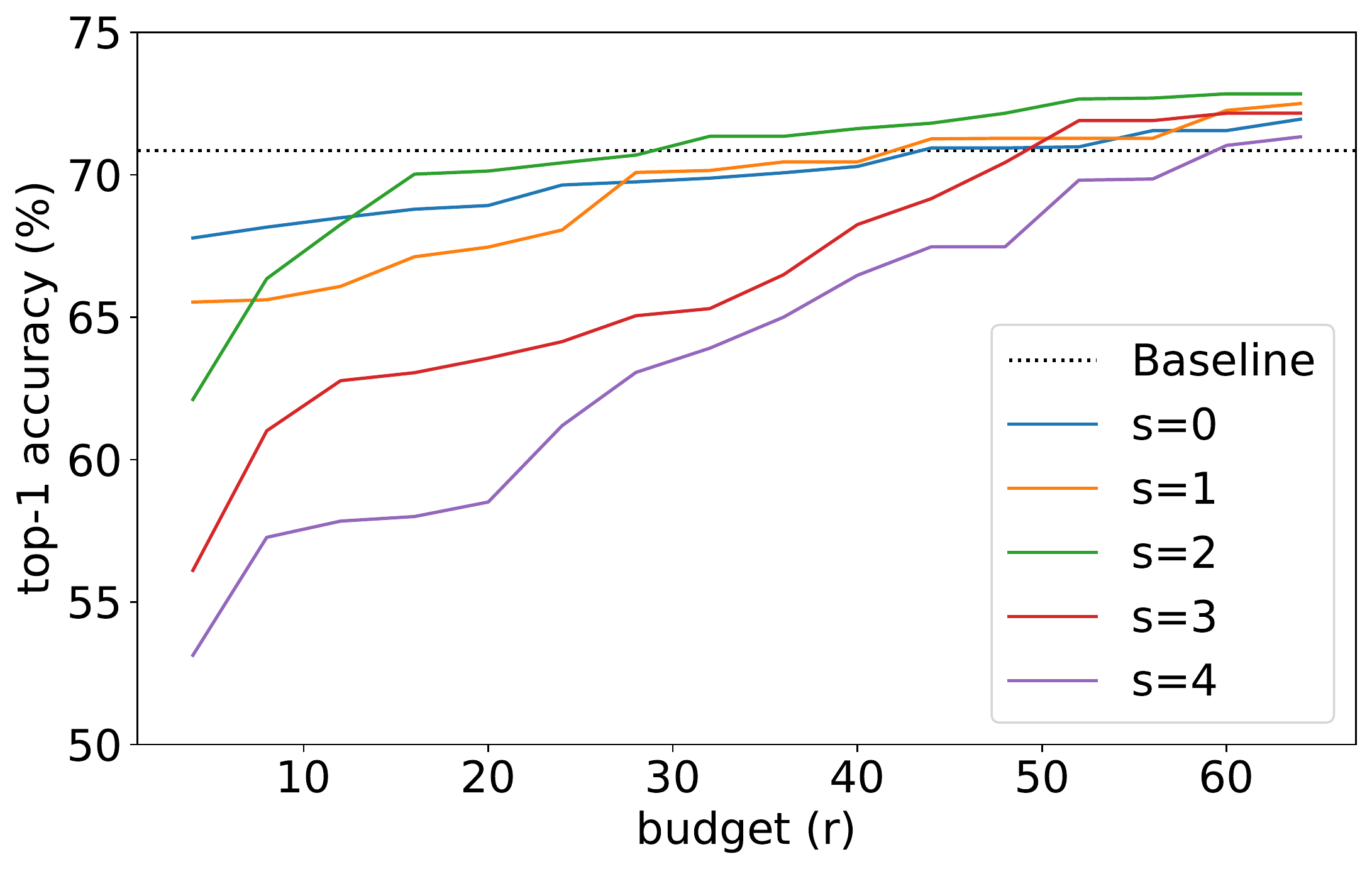}
    \caption{
    Comparison of different minimum exponent $s_{min}$ on CIFAR-100 with VGG-11 architecture.  }
    \label{fig:various_s}
\end{center}
\end{figure}

\vspace{-0.2em}
\subsection{Ablation on Minimum Exponent}
\label{sec:ablation}
We conduct an ablation on minimum exponent $s_{min}$ which adjusts the trade-off between the total number of configurations and minimum resource given a fixed budget. Since a minimum resource of $\eta^{s_{min}}$ is allocated to each configuration, small $s_{min}$ enables to explore more configurations by aggressive early-stopping whereas large $s_{min}$ enables more precise low-fidelity approximation by allocating more resources to each configuration. 

As shown in Figure \ref{fig:various_s}, while $s_{min}=0$ achieves moderate performance in the early phase by exploring order-of-magnitude more configurations, it shows a slow rate of improvement due to imprecise low-fidelity approximation.
On the other hand, it exhibits comparably low performance throughout the optimization process when $s_{min}$ is large because only a small number of configurations can be explored. This result indicates the importance of allocating adequate initial resource and exploring various configurations. While $s_{min}=2$ works generally well throughout the various tasks, datasets and architectures, one might combine MORL with Hyperband which automates the choice of $s_{min}$. We found that Hyperband works well with MORL, showing slightly worse performance than $s_{min}=2$ (72.8 vs 72.6).

\subsection{Integration with Bayesian Optimization}
\label{sec:bayesian}
Bayesian optimization adaptively suggests configuration to evaluate given observations of hyperparameter configuration and its corresponding performance. By incorporating the prior observations into fitting a probabilistic model, it enables to explore more probable candidates compared to random search  \cite{bergstra2011algorithms, chen2018bayesian}. However, vanilla Bayesian optimization typically requires a lot of resources to achieve satisfying performance as it needs to train a full model to get one observation. Previous work \cite{falkner2018bohb} showed that Bayesian optimization can be successfully combined with multi-fidelity optimization to benefit from the advantage of each method. 

\begin{figure}
\begin{center}
    \includegraphics[width=0.44\textwidth]{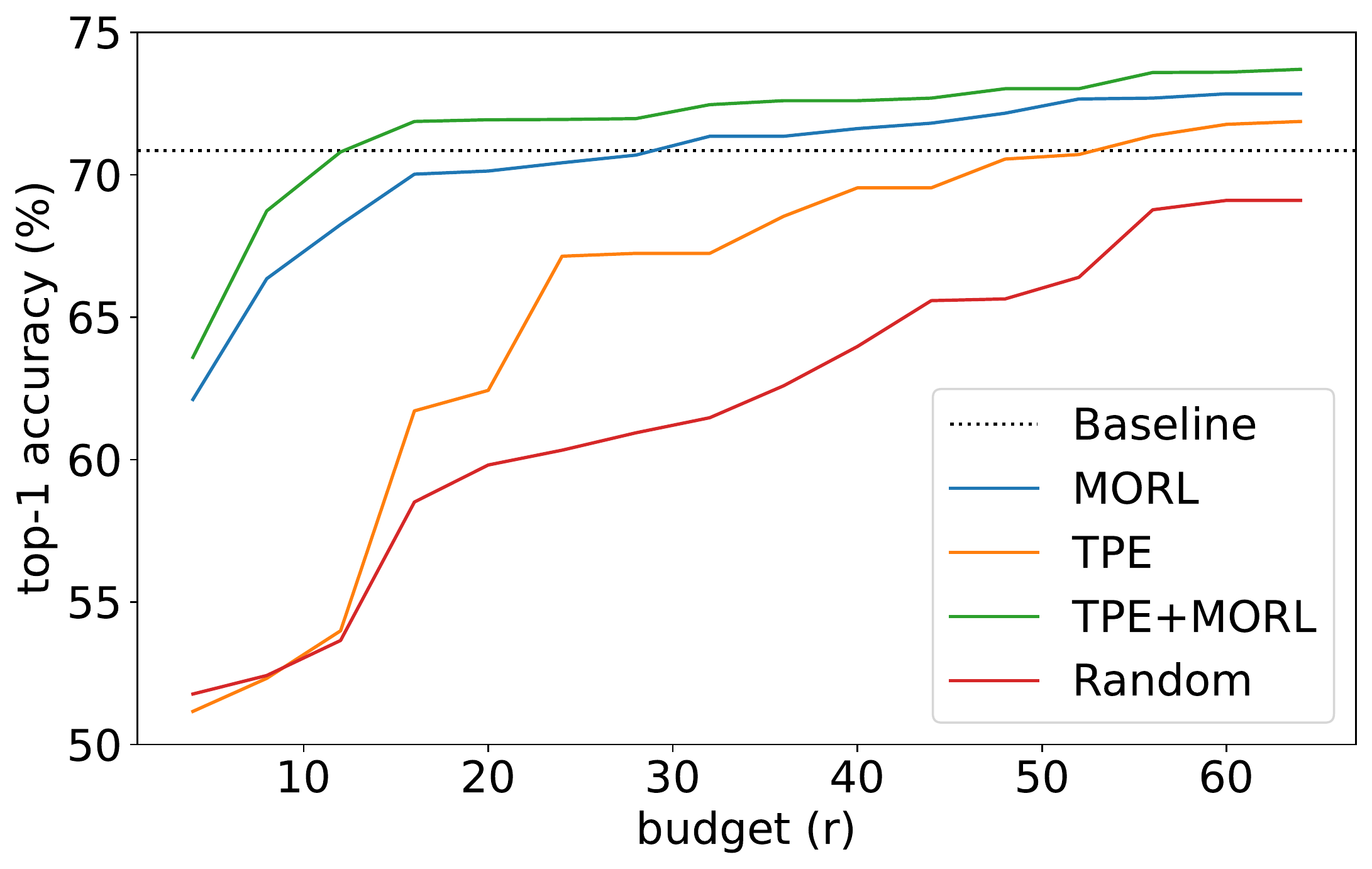}
    \caption{
    Performance of MORL combined with Tree-structured Parzen Estimator (TPE) on CIFAR-100 with VGG-11. MORL is orthogonal to TPE, complementing the limitation of Bayesian optimization by allowing to explore much more configurations.  }
    \label{fig:tpe}
\end{center}
\end{figure}

Among various Bayesian optimization algorithms, we choose Tree-Structured Parzen Estimator (TPE) \cite{bergstra2011algorithms} which models density function based on good and bad observations. It is widely adopted because it is relatively fast and robust to high dimension. We borrow \hj{a} TPE implementation of the popular HPO framework \cite{optuna_2019} with the multivariate option which considers the dependencies among hyperparameters. As shown in Figure \ref{fig:tpe}, TPE exhibits significantly better performance compared to random search as it suggests more plausible configurations. However, it shows limited performance as it only evaluates a small number of configurations. 
When integrating MORL with TPE, it consistently shows the best performance throughout the HPO process. 
It implies that the advantage of MORL is orthogonal to Bayesian optimization methods and they complement each other by exploring more configurations and adaptively suggesting configurations.
We \hj{summarize} top-performing configurations in Table \ref{table:tpe_hyperparam}\hj{. The top-5 high-performing} configurations are \hj{diversely distributed, not similar to each other, demonstrating the importance of exploration in HPO.}

\vspace{-0.2em}
\subsection{Various Learning Rate Schedule}
\label{sec:various_lr}

\vspace{-0.2em}
We finally demonstrate the scalability of MORL with respect to various learning rate schedules. 
Specifically, we further examine MORL with step, cyclical and linear learning rate which are widely used and adopted in popular deep learning framework \cite{NEURIPS2019_9015}. Step learning rate decays the learning rate by a given factor once the epoch reaches the specified milestone, cyclical learning rate \cite{smith2017cyclical} monotonically increases then decreases learning rate within a cycle and linear learning rate linearly decreases learning rate given cycle. 
Each learning rate schedule is condensed to fit each round of promotion and restarted every round. For the step learning rate, we follow the ratio of decaying epoch in section \ref{sec:object-cls}.
Table \ref{table:various_lr} shows the results of various learning rate schedules combined with MORL.
Throughout the all tested learning rate schedules, MORL boosts the performance of the baseline by a considerable margin. Furthermore, the performance gap among varying schedules is within the allowable range. These results show that the advantage of MORL is not restricted to a certain learning rate schedule but can be extended to various schedules.

\begin{table}
\caption{Top-5 configurations of VGG-11 on CIFAR-100 obtained by MORL combined with Bayesian optimization. (LR: learning rate, WD: weight decay, MMT: momentum, BS: batch size).}
\begin{center}
\begin{tabular}{|c|c|c|c|c|c|c|c|}
\hline
 & LR & WD & MMT & BS & Accuracy \\
\hline
Top-1 & 0.0284 & 0.0147 & 0.0596  & 89 & 73.71 \\
Top-2 & 0.0039 & 0.0148 & 0.7976  & 71 & 73.53 \\
Top-3 & 0.0122 & 0.0187 & 0.7371  & 218 & 73.50 \\
Top-4 & 0.0072 & 0.0216 & 0.3322  & 69 & 73.24 \\
Top-5 & 0.0128 & 0.0218 & 0.3417  & 107 & 73.15 \\

\hline
\end{tabular}
\end{center}
\label{table:tpe_hyperparam}
\end{table}

\begin{table}
\caption{Comparison of different learning rate schedules combined with MORL on CIFAR-100 with VGG-16. The effectiveness of MORL is not constrained to a certain learning rate schedule.}
\begin{center}
\begin{tabular}{|c|c|c|c|c|c}
\hline
 & Top-1 Accuracy  \\
\hline
Baseline & 73.58 \\ 
Step LR + MORL & 75.80  \\ 
Linear LR + MORL & 75.89  \\ 
Cyclical LR + MORL & 75.68  \\ 
Cosine annealing LR + MORL & \textbf{76.01}  \\ 
\hline
\end{tabular}
\end{center}
\label{table:various_lr}
\end{table}

\vspace{-0.3em}
\section{Conclusion}
\label{sec:conclusion}
\vspace{-0.2em}
In this work, we present Multi-fidelity Optimization with a Recurring Learning rate (MORL) which enables precise low-fidelity approximation among hyperparameter configurations.
By incorporating \hj{the} learning rate schedule into \hj{the} multi-fidelity optimization process, MORL 
achieves to search outstanding configuration within a practical budget.
Our extensive experiments on a wide range of settings demonstrate the effectiveness of MORL in hyperparameter optimization. While previous works often fail to improve hand-tuned hyperparameters, MORL successfully outperforms human experts by a \hj{significant} margin. Furthermore, we verify that \hj{the proposed method} is orthogonal to Bayesian optimization and applicable to various learning rate schedules. We hope that our work contributes to \hj{automating} hyperparameter tuning process and \hj{pushing} the boundary of \hj{CNN's performance} in various applications.

{\small
\bibliographystyle{ieee_fullname}
\bibliography{wacv}
}

\end{document}